\documentclass[10pt,twocolumn,letterpaper]{article}

\usepackage{iccv}
\usepackage{authblk}
\usepackage{times}
\usepackage{epsfig}
\usepackage{graphicx}
\usepackage{amsmath}
\usepackage{amssymb}
\usepackage{enumitem}
\usepackage{multirow}
\usepackage{booktabs}
\usepackage{marvosym}
\usepackage{indentfirst} 
\usepackage{graphicx}
\usepackage{bbm}
\usepackage[T1]{fontenc}
\usepackage[utf8]{inputenc}
\usepackage[english]{babel}
\usepackage[autostyle, english=american]{csquotes}
\usepackage{subcaption}
\usepackage{ulem}
\usepackage{pifont}

\MakeOuterQuote{"}
\usepackage{bbm}

\def\x{{\boldsymbol x}}

\def\y{{\boldsymbol y}}

\def\DM{{\mathcal D}}

\usepackage[breaklinks=true,bookmarks=false]{hyperref}

\iccvfinalcopy 

\def\iccvPaperID{****} 

\ificcvfinal\fi
\ificcvfinal\fi
\begin{document}
\normalem
\title{Solution for OOD-CV UNICORN Challenge 2024 Object Detection Assistance LLM Counting Ability Improvement}
\author{

Zhouyang Chi$^1$,
 Qingyuan Jiang$^1$,
 Yang Yang$^1$
}

\affil{ 
 $^1$Nanjing University of Science and Technology
}

\maketitle
\setlength{\intextsep}{1pt}
\setlength{\abovecaptionskip}{1.5pt}
\begin{abstract}

 This report provide a detailed description of the method that we explored and proposed in the ECCV OOD-CV UNICORN Challenge 2024, which focusing on the robustness of responses from large language models. The dataset of this competition are OODCA-VQA and SketchyQA. In order to test the robustness of the model. The organizer extended two variants of the dataset OODCV-Counterfactual and Sketchy-Challenging. There are several difficulties with these datasets. Firstly, the Sketchy-Challenging dataset uses some rarer item categories to test the model's generalization ability. Secondly, in the OODCV-Counterfactual dataset, the given problems often have inflection points and computational steps, requiring the model to recognize them during the inference process. In order to address this issue, we propose a simple yet effective approach called Object Detection Assistance Large Language Model(LLM) Counting Ability Improvement(ODAC), which focuses on using the object detection model to assist the LLM. To clarify, our approach contains two main blocks: (1)Object Detection Assistance. (2) Counterfactual Specific prompt. Our approach ranked second in the final test with a score of 0.86.
\end{abstract}
\begin{figure*}
    \centering
    \includegraphics[scale=0.49]{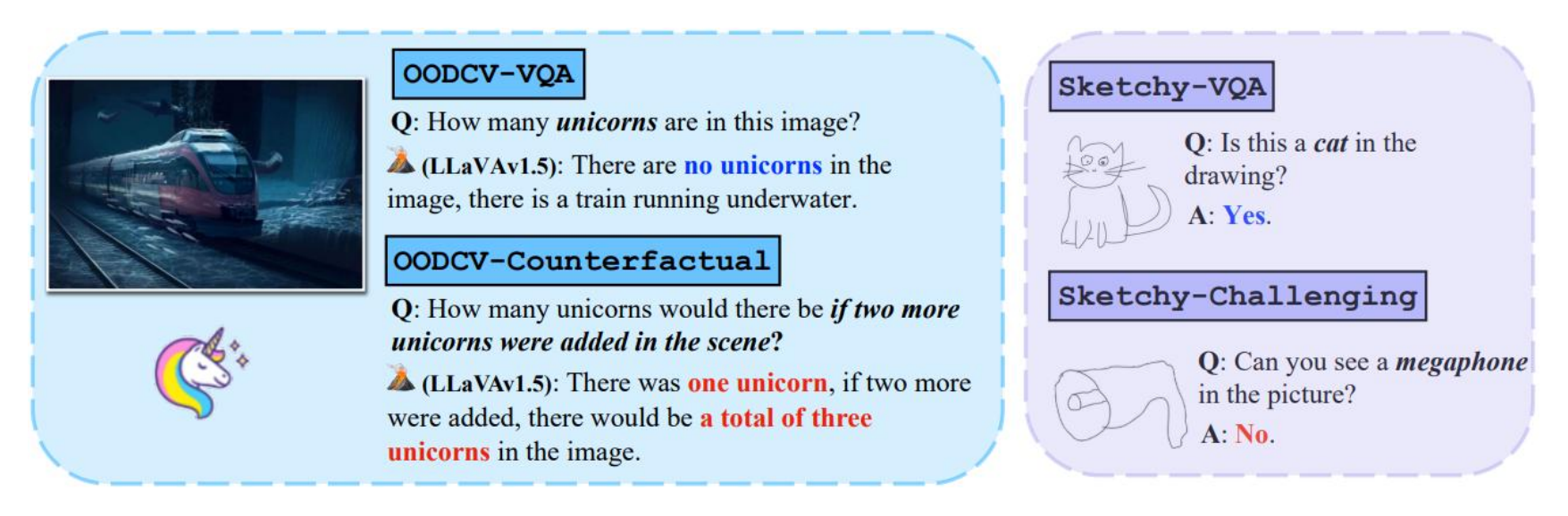}
    \caption{OODCV-VQA, Sketchy-VQA and their variants.}
    \label{fig: dataset}
    \vspace{-10pt}
\end{figure*}
\begin{figure}
    \centering    
    \includegraphics[width=1\columnwidth]{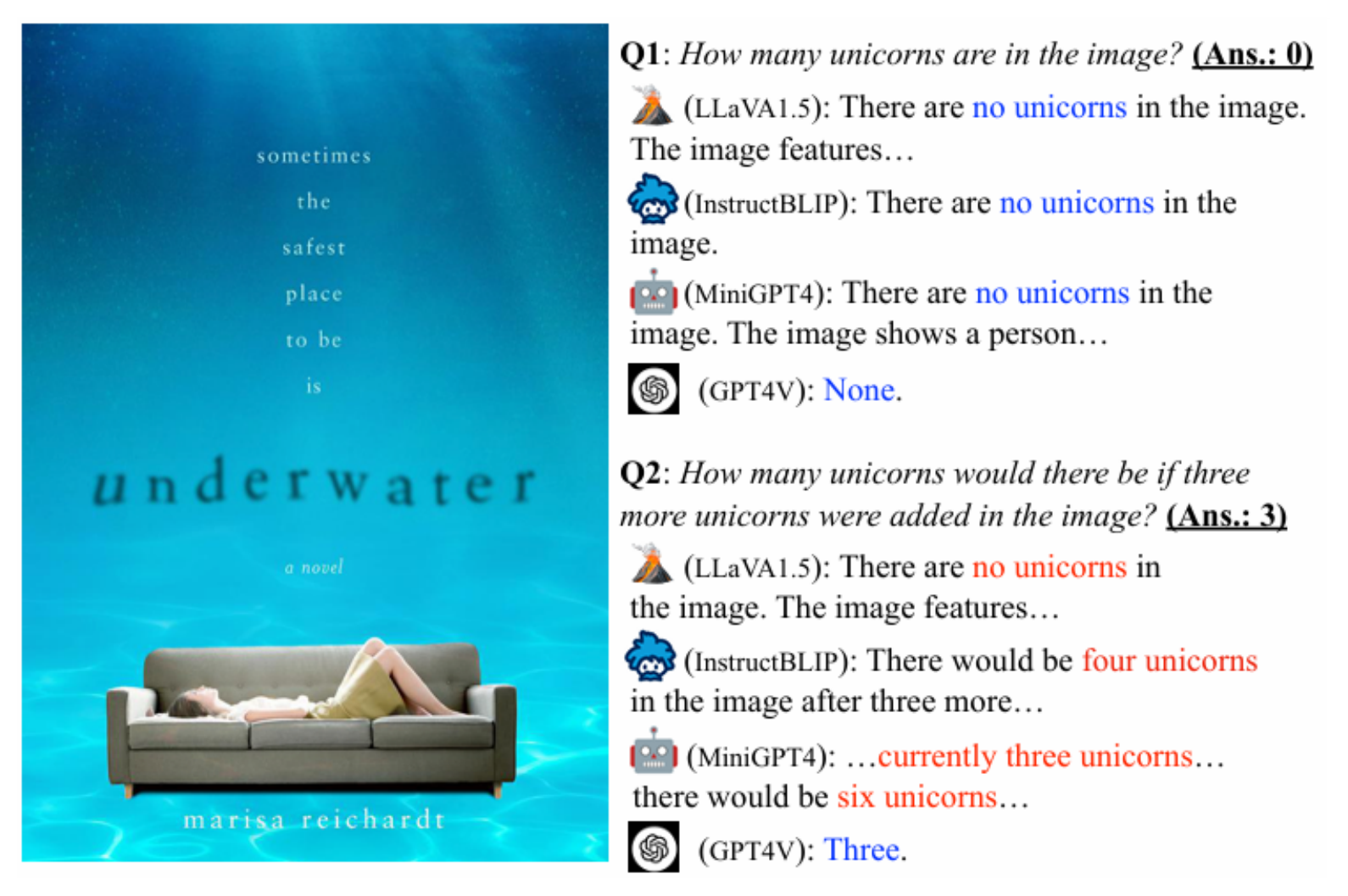}
    \caption{Illusionary answers that occur when a large model encounters counterfactual problems.}
    \label{fig: false}    
    \vspace{-15pt}
\end{figure}
\section{Introduction}

In recent years, multimodal learning\cite{yang2019comprehensive, yang2019semi, yang2019semi_2} has gradually become one of the research hotspots in the fields of machine learning and data mining. It has been successfully applied in various real-world scenarios, such as multimedia search\cite{yang2021rethinking}, multilingual processing\cite{DOMFN:conf/mm/0074ZGGZ22}, video segementation\cite{TGVSS:conf/aaai/YangHGXX23}, image captioning\cite{yang2022exploiting,NAIC:conf/aaai/FuSZY24} and recommendation\cite{SMHUI:conf/wsdm/ZhuCLYYX20}. With the emergence of multimodal large language models, many multimodal problems have been properly solved, but with it comes some hidden problems. The answer of the large language model is always very uncertain and not robust enough\cite{Unicorn:journals/corr/abs-2311-16101}.\par

In order to improve the response robustness of the large language model, the organizers published four datasets as shown in Figure \ref{fig: dataset}, OODCV-VQA, Sketchy-VQA and its variation. These datasets are interesting and important. It includes some common categories as well as uncommon categories to test the OOD recognition ability of the model. And unlike conventional visual semantic question answering tasks, the text has significant interference, such as inflection or reasoning parts. In this competition, we used a pretrained large multimodal model: InstructBLIP\cite{InstructBLIP:conf/nips/Dai0LTZW0FH23} as our baseline.\par
According to Figure \ref{fig: dataset}, pretrained large models tend to be more accurate in answering common items like cat, while they are prone to errors in answering some less common items like megaphone. What's more, consider Figure \ref{fig: false}, when a large language model is faced with common problems, it can provide more appropriate answers in most cases, but when encountering counterfactual problems, the large model may encounter problems of illusion and reasoning errors\cite{Hall:journals/corr/abs-2311-05232}. So we can see that the answer results of the current large model are not robust enough and even a slight disturbance can destroy its reasoning ability to a certain extent. \par
\begin{figure*}
    \centering
    \includegraphics[scale=0.49]{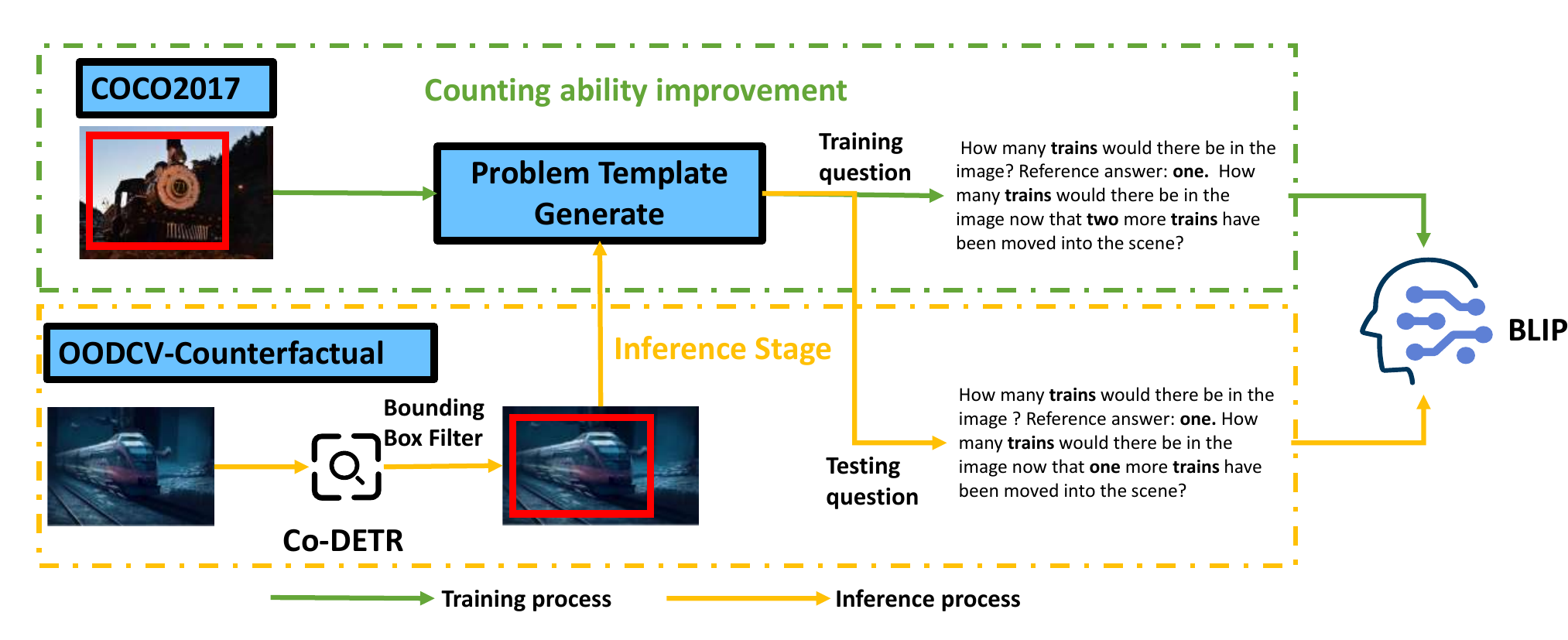}
    \caption{Outline of our proposed method. Pseudo Labels generated from object detection model and well designed counterfactual specific prompt.}
    \label{fig: architecture}
    \vspace{-10pt}
\end{figure*}

To address these challenges, we made many attempts and ultimately achieved success in helping large models improve their reasoning and counting abilities which means that we mainly helped large models achieve better results on OODCV-VQA and its variants counterfactual dataset. So, we propose Object Detection Assistance LLM Counting Abiltity Improvement, as shown in Figure \ref{fig: architecture}. Firstly, in order to improve the counting ability of large lange model, we choose to use object detection models to generate pseudo labels to help improve the counting ability of the model. Secondly, in order to enhance the computational reasoning ability of the model, we propose the Counterfactual Specific prompt, which enhance the text input.

Our contributions can be summarized as follows:
\begin{itemize}
[itemsep=0pt,parsep=0pt,topsep=0pt,partopsep=0pt,leftmargin=*]
    \setlength{\itemindent}{1.3em}
    \item When a large model answers counterfactual questions, it may experience hallucinations, leading to a decrease in performance. Counterfactual Specific prompt design can improve the model's counting ability and avoid this problem.
    \item We found that professional models still outperform ordinary multimodal large models in a single domain. Therefore, we use professional small models to assist large models in answering questions.
\end{itemize}

\section{Method}
\subsection{Base Model}
In our work, we take InstructBLIP as our base model. And we use the Co-DETR\cite{DETR:conf/iccv/ZongS023} object detection model to generate the pseudo label. 
\vspace{-10pt}
\subsubsection{InstructBLIP} 
\setlength{\parindent}{2em}  InstructBLIP is a visual instruction tuned version of BLIP-2\cite{BLIP:conf/icml/0001LXH22,BLIP2:conf/icml/0008LSH23}. InstructBLIP is a vision-language instruction tuning framework that enables general-purpose models to solve a wide range of visual language tasks through a unified natural language interface. InstructBLIP uses a diverse set of instruction data to train a multimodal LLM. Specifically, initialize training with a pre-trained BLIP-2 model consisting of an image encoder, an LLM, and a Query Transformer (Q-Former) to bridge the two.  During instruction tuning, finetune the Q-Former while keeping the image encoder and LLM frozen. Additionally, they introduce an instruction-aware Query Transformer, which extracts informative features tailored to the given instruction.\par
\vspace{-10pt}
\subsubsection{Co-DETR}
    Co-DETR is a novel collaborative hybrid assignments training scheme and learn more efficient and effective DETR-based detectors from versatile label assignment manners. The proposed training scheme can easily enhance the encoder's learning ability in end-to-end detectors by training multiple parallel auxiliary heads supervised by one-to-many label assignments. What's more, author conduct extra customized positive queries by extracting the positive coordinates from these auxiliary heads to improve attention learning of the decoder.
\subsection{Object Detection Assistance Improvement}
In order to address the issue of hallucinations and decreased reasoning ability caused by counterfactual problems in large models, we propose Object Detection Assistance LLM Counting Ability Improvement. It consists of two parts, namely Obecjt Detection Assistance Improvement and Counterfactual Specific prompt. Specifically, we sampled from the COCO dataset\cite{COCO:conf/eccv/LinMBHPRDZ14} to form our fine-tuning\cite{SDSU:journals/corr/abs-2303-15647} dataset and used the two methods mentioned above to fine tune the InstructBLIP model to obtain the final results.
\vspace{-10pt}
\subsubsection{Object Detection Assistance}
We first used the object detection model Co-DETR, which currently performs the best on the COCO dataset to generate the corresponding pseudo labels. For example, 'How many bicycles are there in the picture?', the corresponding pseudo label is the number of bicycles detected in the picture by Co-DETR. Then we sample some data from COCO dataset as our finetuning dataset. The corresponding problems are randomly generated, similar to OODCV and OODCV-Challenge problems, and include the following types: yesno problem, digit problem, yesno problem with inflection, and digit problem with addition and subtraction. In order to improve the performance of pseudo labels, we perform appropriate filtering on the detection results, such as filtering detection boxes with aspect ratios below a certain threshold and confidence levels below a certain threshold which is 0.8 in our code.\par
\vspace{-10pt}
\subsubsection{ Counterfactual Specific prompt}
For different datasets, we designed different prompts for targeted fine-tuning. For the OODCV dataset, during training, we combined the generated yesno problems, digit problems, and labels of COCO. During testing, we combined the original problems from OODCV dataset with the pseudo labels. For the OOODCV-Challenge dataset, we combine the yesno problem with inflection, and digit problem with addition and subtraction. \par
We define the testing dataset $\DM=\{\x_i^{(t)},\x_i^{(i)},\y_i\}_{i=1}^n$ and $\DM_{cf}=\{\x_i^{(cf-t)},\x_i^{(i)},\y_i^{(cf)}\}_{i=1}^n$, represent the text modality, the image modality, and the labels. And the COCO fine-tuning dataset is defined by $\DM_{coco}=\{\x_i^{(coco-t)},\x_i^{(coco-i)},\y_i^{(coco)}\}_{i=1}^m$. $\varphi(\cdot)$ denote the large-language model (LLM). $\psi(\cdot)$ denote the object detection model. $\{\y^{(p)}_i\}_{i=1}^{n}$ means the pseudo labels.\\
We first use a pre-trained Co-DETR object detection model to perform object detection on the image modality of the test data and generate pseudo labels. 
\begin{equation}
    \label{equ:pseudo_label}
    \{\y^{(p)}_i\}_{i=1}^{n}=\psi(\x_i^{(i)})
\end{equation}
For the OODCV dataset, the specific prompt design is as follows.\\
\textbf{Training-yesno}: 'Is there a XXXX in the image?' or 'Would there be a XXXX in the image once the XXXX has been removed from the scene?'\\
\textbf{Training-digit}: 'How mant XXXXs would there be in the image? Reference answer: ZZZZ.'\\
For the OODCV-Counterfactual dataset, the specific prompt design is as follows.\\
\textbf{Training-yesno}: 'Is there a XXXX in the image?' or 'Would there be a XXXX in the image once the XXXX has been removed from the scene?'\\
\textbf{Training-digit}: 'How many XXXXs would there be in the image? Reference answer: ZZZZ. How many XXXXs would there be in the image now that YYYY more XXXXs have been moved into the scene? ' or 'How many XXXXs would there be in the image? Reference answer: ZZZZ. How many XXXXs would there be in the image if someone added YYYY more XXXXs in the picture? '\par
XXXX in the text represents category information which is same to original problem. YYYY is a randomly generated number from 0 to 5 and we will add or subtract that number from the original labels of COCO or pseudo label. ZZZZ is always the label of COCO and replace with pseudo labels during testing. During the training, we also found that when fine-tuning the model, it is best to use 'ten' instead of '10' for labels. This can be achieved directly by calling the num2words package. Finally, labels can be represented using the following formula. The ratio of yesno problems and digit type problems is also a key factor in improving model performance, as a yesno type problem can lead to overfitting of the model. Therefore, it is necessary to reduce the amount of data for this type of problem. Here, we have adopted a 1:100 ratio.
\begin{equation}
    y_i^{(train)}=num2words(y_i^{(coco)}\pm \mathrm{YYYY})
\end{equation}
\begin{equation}
     y_i^{(test)}=num2words(y_i^{(p)}\pm \mathrm{YYYY})
\end{equation}
\par
After that, we can start fine-tuning our model.\par
\subsection{Discussion}
This is all the exploration we have made in the competition. Both methods have achieved significant success and it has improved the robustness of the model. But there has not been much improvement in the generalization ability of large models for rare classes in OOD detection. This is also a future research direction.

\section{Experiment}
\textbf{Dataset.} Our method is fine-tuned on the pretrained models of InstructBLIP using official datasets. And we follow the official way to split the training and validation sets.

\textbf{Metric.} Even if the entire task is a model generated task, the corresponding answers can be summarized by one word, such as' yes' corresponding to 1 and 'no' corresponding to 0. So after processing the sentences generated by the model, we can use ACC as our evaluation metric.

\textbf{Implementation Detail.} 
Our approach is founded on fine-tune principles and makes use of the pre-trained model made available on the official  InstructBLIP\footnote{\label{myfootnote1}$https://huggingface.co/collections/Salesforce/instructblip-models-65242fd466ebe051985ec483$} and X$^2$-VLM website
\footnote{\label{myfootnote2}$https://github.com/Sense-X/Co-DETR$}. We use SGD optimizer and set the learning rate at 1e-5 and weight decay 2e-4 to fine tune the model. All experiments are deployed on one
NVIDIA A100 and nearly need 60G memory in GPU.

\begin{table}[htp]
\vspace{1.0em}
    \centering
    \begin{tabular}{cccccc}
    \toprule
    Method & Acc \\
    \hline
    $\textbf{BLIP-Large}$ & 0.683 \\
    \textbf{Qwen-VL-Chat\cite{Qwen:journals/corr/abs-2308-12966}}& 0.813\\
    \textbf{InstructBLIP-FlanT5-XL} & 0.724 \\
    \textbf{Co-DETR} & 0.819\\
    \toprule
    $\textbf{BLIP-Large}^*$& 0.767 &  \\
    $\textbf{Qwen-VL-Chat}^*$& 0.848\\
    $\textbf{InstructBLIP-FlanT5-XL}^*$& \textbf{0.862} \\
    \toprule
    \end{tabular}
\caption{Comparison method. * means the method use the ODAC which we proposed.}
\label{tab: compare}
\end{table}

\textbf{Comparison Methods Result.} Table \ref{tab: compare} shows the average accuracy score performance on four datasets. The accuracy of Co-DETR is calculated on the OODCV-VQA dataset and OODCV counterfactual dataset.\par
Note that regardless of the size of the pretrained model from BLIP to Qwen, there is a varying degree of improvement after adding our method. This proves the universality of our method. And we can find that the performance of object detection models can even exceed that of large models, so improving the robustness of the model's answers through professional models is a feasible approach.

\textbf{Ablation Study.} To analyze the contribution of the pseudo label and prompt in our method, we conduct more ablation studies in this competition. The Acc score after adding different prompts is demonstrated in Table \ref{tab: abl}. From the table, we can clearly see that after adding prompt and the pseudo label, the Acc value has a certain improvement.
\begin{table}[htp]
\vspace{1.0em}
    \centering
    \begin{tabular}{cccccc}
    \toprule
    Dataset &ODA&CSP& Acc \\
    \hline
    \multirow{4}*{OODCV-VQA}&\ding{56} & \ding{56} & 0.714 \\
     &\ding{52} & \ding{56} & 0.782\\
     &\ding{56} & \ding{52} & 0.736\\
     &\ding{52} & \ding{52} & 0.802\\
    \toprule
    \multirow{4}*{OODCV-Counterfactual}&\ding{56} & \ding{56} & 0.480 \\
     &\ding{52} & \ding{56} & 0.691\\
     &\ding{56} & \ding{52} & 0.558\\
     &\ding{52} & \ding{52} & 0.814 \\
    \toprule
    \end{tabular}
\caption{Ablation experiment. ODA and SCP denote Object Detection Assistance and Counterfactual Specifical Prompt.} 
\label{tab: abl}
\end{table}
\begin{table}[htp]
\vspace{1.0em}
    \centering
    \begin{tabular}{cccccc}
    \toprule
    Hyperparameters &Value& Yesno Acc & Digit Acc\\
    \hline
    \multirow{4}*{Confidence threshold}&0.10 & 0.723& 0.312 \\
     &0.70 & 0.923& 0.752\\
     &0.85 & 0.997 & 0.796\\
     &1.00 & 0.813 & 0.674\\
    \toprule
    \multirow{3}*{Problem ratio}&1:1 & 1.00 & 0.343\\
     &1:10 &  1.00 & 0.625\\
     &1:100 & 0.997 & 0.796\\
    \toprule
    \end{tabular}
\caption{This table shows the sensitivity performance of our method to hyperparameters on dataset OODCV-Counterfactual. Yesno Acc means the accuracy of yes/no problem. Digit Acc means the accuracy of digit problem.} 
\label{tab: hyper}
\end{table}
\textbf{Hyperparameter research.} In order to investigate the impact of hyperparameters on our method, we conducted experiments on the detection confidence filtering threshold of the object detection model, as well as the proportional parameters of the yesno problem and the digit problem. The final results are shown in the table \ref{tab: hyper}.
\section{Conclusion}

This report summarized our solution for the OOD-CV UNICORN Challenge 2024. Our approach was based on the object detection assistance and counterfactual specifical prompt which help the model improve its target counting ability in the face of complex language conditions and enhance the robustness of the LLM model's response.

{\small
\bibliographystyle{ieee_fullname}
\bibliography{main}
}
\end{document}